\documentclass[runningheads]{llncs}
\usepackage{graphicx}
\usepackage[linesnumbered,ruled,vlined]{algorithm2e}
\usepackage{booktabs}
\usepackage{siunitx}
\usepackage{multirow}
\usepackage{rotating}

\begin{document}
\title{Knowledge Graphs for Enhancing Large Language Models in Entity Disambiguation}
\author{Gerard Pons\inst{1}\and Besim Bilalli\inst{1}\and Anna Queralt
 \inst{1}}
\authorrunning{G. Pons et al.}
%\institute{UPC}
\institute{Universitat Politècnica de Catalunya, UPC-BarcelonaTech \email{\{gerard.pons.recasens,besim.bialli,anna.queralt\}@upc.edu}}
\maketitle              % typeset the header of the contribution
\begin{abstract}
Recent advances in Large Language Models (LLMs) have positioned them as a prominent solution for Natural Language Processing tasks. Notably, they can approach these problems in a zero or few-shot manner, thereby eliminating the need for training or fine-tuning task-specific models. However, LLMs face some challenges, including hallucination and the presence of outdated knowledge or missing information from specific domains in the training data. These problems cannot be easily solved by retraining the models with new data as it is a time-consuming and expensive process. To mitigate these issues, Knowledge Graphs (KGs) have been proposed as a structured external source of information to enrich LLMs. With this idea, in this work we use KGs to enhance LLMs for zero-shot Entity Disambiguation (ED). For that purpose, we leverage the hierarchical representation of the entities' classes in a KG to gradually prune the candidate space as well as the entities’ descriptions to enrich the input prompt with additional factual knowledge. Our evaluation on popular ED datasets shows that the proposed method outperforms non-enhanced and description-only enhanced LLMs, and has a higher degree of adaptability than task-specific models. Furthermore, we conduct an error analysis and discuss the impact of the leveraged KG's semantic expressivity on the ED performance.
\keywords{Knowledge Graphs  \and Entity Disambiguation \and Large Language Models.}
\end{abstract}

\section{Introduction}
\label{sec:introduction}
The association of textual mentions in a document to the entities they refer to in a Knowledge Graph (KG) is crucial for many Natural Language Processing (NLP) applications, such as question answering or information retrieval. This task is known as Entity Linking (EL), and it is a fundamental  step in the transformation of unstructured text into structured knowledge. EL is usually performed as a pipeline with three different steps. The first one is Mention Detection, which detects the text spans that could possibly be linked to entities. It is followed by the Candidate Generation stage, which selects for each mention the top k entities from the KG that could refer to it, usually based on precomputed probability distributions from entity-mention hyperlink pairs. Finally in the Entity Disambiguation (ED) step, a final entity is selected from the previously generated set.

Usually, the ED problem is tackled by designing and training task-specific models with large amounts of data (e.g., Wikipedia dumps)~\cite{genre20,refined22}. In recent years, language models have been used for this task by making use of the mention’s context in the document to disambiguate between the possible solutions~\cite{titov18,genre20,extended22}. Additionally, some approaches incorporate the candidates’ descriptions~\cite{descriptions19}, classes~\cite{types20} (e.g., the categories they are tagged with in Wikipedia) or both~\cite{refined22}  in the model’s input, by generating encodings for these text items. The addition of this knowledge allows zero-shot ED, enabling the models to classify entities that may have not been seen during training time.

Lately, new advances in Large Language Models (LLMs) such as GPT-3~\cite{gpt320}, GPT-4~\cite{gpt423} or LLaMA-2~\cite{llama223}, have demonstrated remarkable performance in numerous NLP problems~\cite{llmspower23}. Given their large-scale and diverse training corpus, they are good candidates to perform tasks, even zero-shot ones, where general knowledge is needed for language processing, such as ED~\cite{chatel24}. However, these LLMs still face some challenges, such as hallucination (i.e., the generation of statements that are factually incorrect)~\cite{hallucination23}, and the lack of knowledge about concepts outside their training corpus. To mitigate these issues, the use of KGs to enhance LLMs has been proposed to address different problems~\cite{roadmap24}. There exist a large variety of KGs, storing information which can be encyclopedic (e.g., DBpedia~\cite{dbpedia} or YAGO~\cite{yago}, which extract information from Wikipedia), commonsense knowledge (e.g., ConceptNet~\cite{conceptnet}, with information such as $\langle house,has,door\rangle$ or $\langle bed,usedFor,sleep\rangle$) or domain specific~\cite{domainkg}. The explicit and structured knowledge they contain can be used to enhance the performance of LLMs, by leveraging it either during pre-training by enriching the training data~\cite{kgpretraining}, or during the inference stage~\cite{kginferencemindmap,kginferenceprompting,kginferencethink}. Following the nomenclature proposed in~\cite{roadmap24}, in this work we focus on KG-enhanced LLM inference, and apply it to the ED task. Our approach takes advantage of KGs to avoid re-training the LLM, and improves the effectiveness of zero-shot LLM approaches for ED. 

%In this work we focus on the latter, thus eliminating the need for training or fine-tuning task-specific models. Concretely, we show how KGs can be used to improve the effectiveness of zero-shot LLM approaches for ED, which to the best of our knowledge has not been explored before.
%
\begin{figure}
\centering
\includegraphics[width=\textwidth]{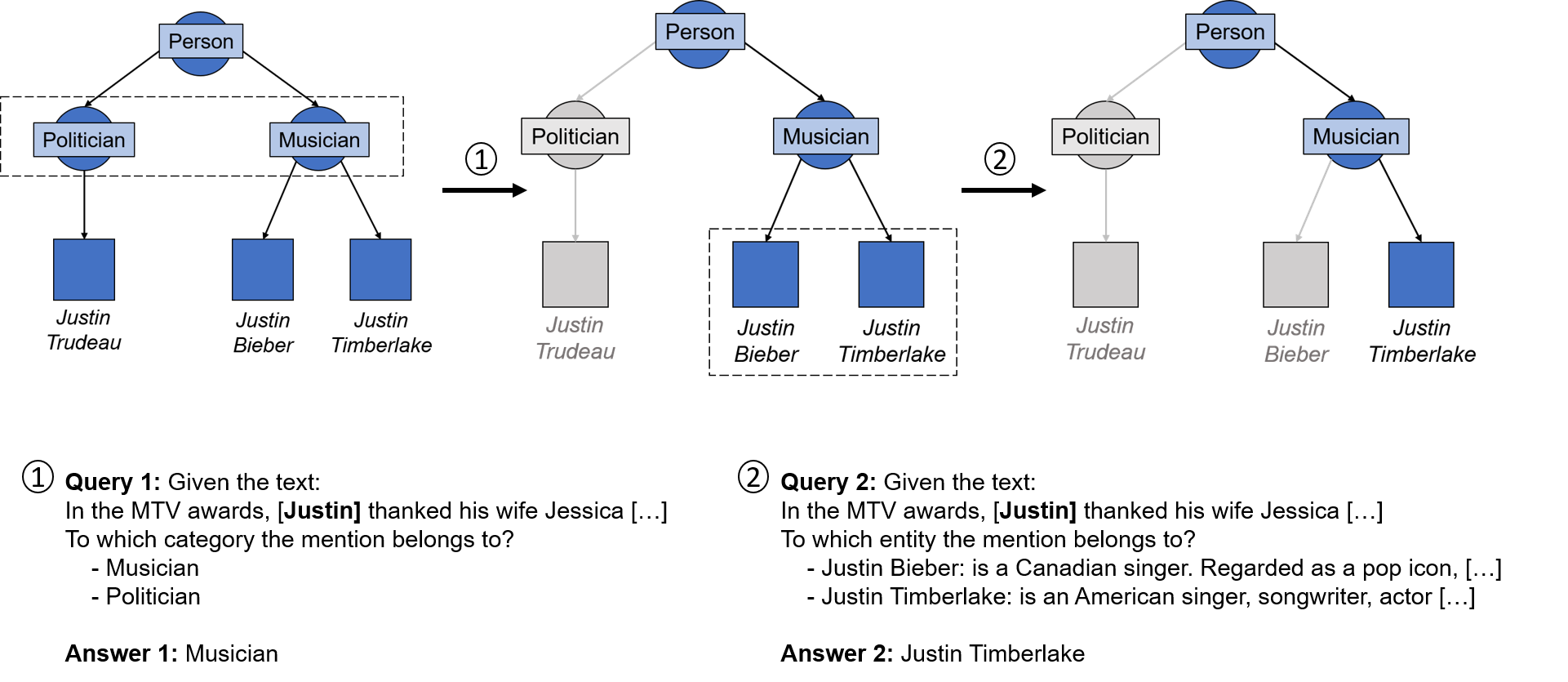}
\caption{Overview of the two steps of our approach.} \label{fig1}
\end{figure}

Solving the ED problem using only LLMs would require to instruct them to choose one of the entities from the candidate set given the document containing the mention. Instead, we propose to extract the candidates' class taxonomy from a KG and use it to guide the disambiguation. For example, taking Query 1 in Figure~\ref{fig1}, given `MTV awards' appearing in the context, the entity `Justin' is more likely to be a \textit{Musician} than a \textit{Politician}. Thus, we can use this context to guide the LLM by eliminating invalid solutions such as `Justin Trudeau', rather than letting the LLM directly predict the entity. Moreover, when all the remaining candidates fall directly under the same class, we retrieve the candidates’ descriptions from a Knowledge Base (KB), such as Wikipedia, and append them to the disambiguation prompt (see Figure~\ref{fig1}, query 2). With this Retrieval Augmented Generation (RAG)~\cite{rag} stage, we provide the LLM with reliable information, reducing hallucination and enabling the LLM to perform predictions over new or unusual entities which may not have been present in the training corpus.  

% Therefore, we assume that given the context and the mention, it is easier for the LLM to infer the mention’s class (e.g., in the example above, given ‘MTV awards’ appearing in the context, the entity is more likely to be a \textit{Musician} than a \textit{Politician}) than to directly predict the underlying entity it refers to.

Therefore, our contributions are as follows:
\begin{itemize}
    \item We present a method to enhance LLMs in the ED task by leveraging the candidate entity class taxonomies available in KGs. Moreover, we also augment the prompt with the entity descriptions, in order to allow the disambiguation of unseen or difficult entities.
    \item We evaluate the method against non-enhanced LLMs, description-only enhanced LLMs and a task-specific model by using ten ED datasets. The results show that our approach improves the disambiguation capabilities of LLMs and has a higher degree of adaptability to different domains than the task-specific model. 
    \item We discuss how using KGs with different levels of semantic expressivity (e.g., YAGO and DBpedia) affects the proposed pruning algorithm, by studying both the ED results and the algorithm's performance.
    \item We study and classify the cases in which our method fails to correctly disambiguate the mention and provide insights on the possible improvements.
\end{itemize}

The remainder of the paper is structured as follows. In Section~\ref{sec:related_work} we discuss the Related Work. In Section~\ref{sec:methodology} we formalize the problem and present the proposed approach. In Section~\ref{sec:experiments} we describe the different experiments, discuss the results, and conduct an error analysis. Finally, in Section~\ref{sec:conclusions} we provide our conclusions and ideas for future work.
\section{Related Work}
\label{sec:related_work}
This section begins with an overview of different ED methods which leverage external information to improve their predictions. Then, the recently emerged concept of KG-enhanced LLMs is introduced, enumerating some of the proposed methods for tackling NLP tasks.
\subsection{Knowledge-augmented ED}
Various ED approaches use model architectures that leverage the mention, its surrounding context and candidate entities to generate a solution~\cite{titov18,genre20,extended22}. However, some recent works incorporate additional knowledge to the model’s input in order to improve the disambiguation of entities which are not present in the training dataset. This extra information is usually gathered from online sources (e.g., Wikipedia and Wikia) and provided in the form of entity descriptions~\cite{descriptions19}, entity types~\cite{types20} or both~\cite{refined22}. Additionally, some works leverage the structured information contained in KGs to enhance the model’s performance. In~\cite{kgzeshel}, information about the entity types from DBpedia and knowledge graph embeddings extracted from Wikipedia's graph structure are incorporated into the model’s input. In~\cite{edkb}, KG triples are used to train a component of the model's architecture which predicts the existence of facts between mentions in a given document. The result of this prediction is used as input for the final model, which also leverages entity types and descriptions. Finally in~\cite{verbalize}, the triples from the KG are verbalized and appended to the input sentence before being fed to the model. 

These knowledge-augmented ED approaches incorporate the additional information to their model’s input, which are mainly built by leveraging LLMs such as BERT~\cite{bert}, RoBERTa~\cite{roberta} or BART~\cite{bart}, and need to be trained or fine-tuned with large amounts of data (e.g., Wikipedia dumps with millions of entities). In contrast, in our approach we rely on the new generation of generative LLMs (e.g., GPT-3~\cite{gpt320}, GPT-4~\cite{gpt423} or LLaMA-2~\cite{llama223}), and solve the ED task by prompting the LLMs in a zero-shot manner without needing to train a task-specific model. This approach has also been explored in~\cite{chatel24}, where the document's context and the inherent knowledge from the LLM are enriched with the entity descriptions, following a RAG approach~\cite{rag}. RAG has been shown to be useful for incorporating new or relevant information to LLMs, and it has also been leveraged in a specific step of our proposal. However, our main focus is on the usage of KGs to obtain the class hierarchy for the candidate entities, which allows our method to solve the ED task by guiding the LLM to the correct answer (see Section~\ref{sec:method}).  

% This approach has been explored in~\cite{chatel24}, where first the LLM is prompted to specify what the mention in the document refers to, and then based on this response the LLM is asked to select the answer among the candidate entities, which are enriched with their description following a RAG approach. RAG has been shown to be useful for incorporating new or relevant information to LLMs, and it has also been used in our proposal. However, we additionally leverage a KG to obtain the class hierarchy and solve the ED task by gradually pruning the candidate space (see Section~\ref{sec:method}).

\subsection{KG-enhanced LLMs}

LLMs can be used to solve a wide range of tasks, not just ED. However, as introduced in Section~\ref{sec:introduction}, LLMs suffer from problems such as hallucination, which can be accentuated if the information requested is outdated or not present in the training data. Retraining LLMs to incorporate this missing knowledge is expensive and time-consuming, and fine-tuning them could lead to problems such as catastrophically forgetting (i.e., the LLMs' tendency to lose previously obtained knowledge when being fine-tuned with new data)~\cite{forget}. To solve these issues, KGs can be used as a source of additional structured information in different NLP tasks. In particular, information from a KG can be added to the prompt fed to the LLMs, a technique coined as KG Prompting~\cite{roadmap24}, which has already been explored for Question Answering. In~\cite{kginferencemindmap}, the approach starts by identifying the entities in the question, and then the KG is queried to build subgraphs including them. After that, the LLM is prompted to comprehend and aggregate the subgraphs, and based on the consolidated result it is asked to reason over it and provide the answer. Similarly in~\cite{kginferencethink}, the LLM generates these subgraphs by iteratively exploring a KG to create a reasoning path over it. In each iteration, if the LLM believes that has enough information, an answer is provided. Otherwise, it is prompted to continue to traverse the graph, adding the most promising relation to the existing reasoning path each time. Finally in~\cite{kginferenceprompting}, the entities are also first extracted from the question, which are then used to retrieve the triples they participate in within the KG. Then, the triples are verbalized and appended to the prompt as context, which is fed to the LLM to obtain the answer.  

In our approach, however, we solve a different task, ED, and we rely on the KG's ontology rather than on the annotated instances, guiding the disambiguation of the entities using the class hierarchy. 
\section{ED with KG-enhanced LLMs}
In this section we lay out the formulation of the problem to be solved and describe the two different steps of the proposed method.
\label{sec:methodology}
\subsection{Problem Formulation}
Let $C = \{e_{1},e_2,...e_{|C|}\}$ be a set of $k$ candidate entities belonging to a KG, containing a class hierarchy in which the entities are annotated, and $m$ be a mention in a document $d$. The objective of ED is to assign to $m$ the entity $e$ it refers to, such that $e \in C$. 

% Given a mention $m$ in a document $d$, the objective of ED is to assign to $m$ the entity $e$ it refers to from a previously computed set of $k$ candidate entities $C = \{e_{1},e_2,...e_{|C|}\}$ (see Section~\ref{sec:settings}). These candidates belong to a KG, where they are annotated under a hierarchical taxonomy of classes and that the entities' descriptions can also be retrieved. To select the entity $e$, the method and the algorithm presented in the next subsections have been developed.
\subsection{Method}
Our proposed method for the disambiguation of the mention can be divided in two steps. First, a subgraph is generated containing the candidate entities together with their taxonomy of classes. Then, a pruning algorithm is applied to iteratively discard  the candidate entities until there is only one left in the subgraph, which will be the solution (see Figure~\ref{fig1}). The implementation can be found in the Supplemental Material.

\begin{figure}
\includegraphics[width=\textwidth]{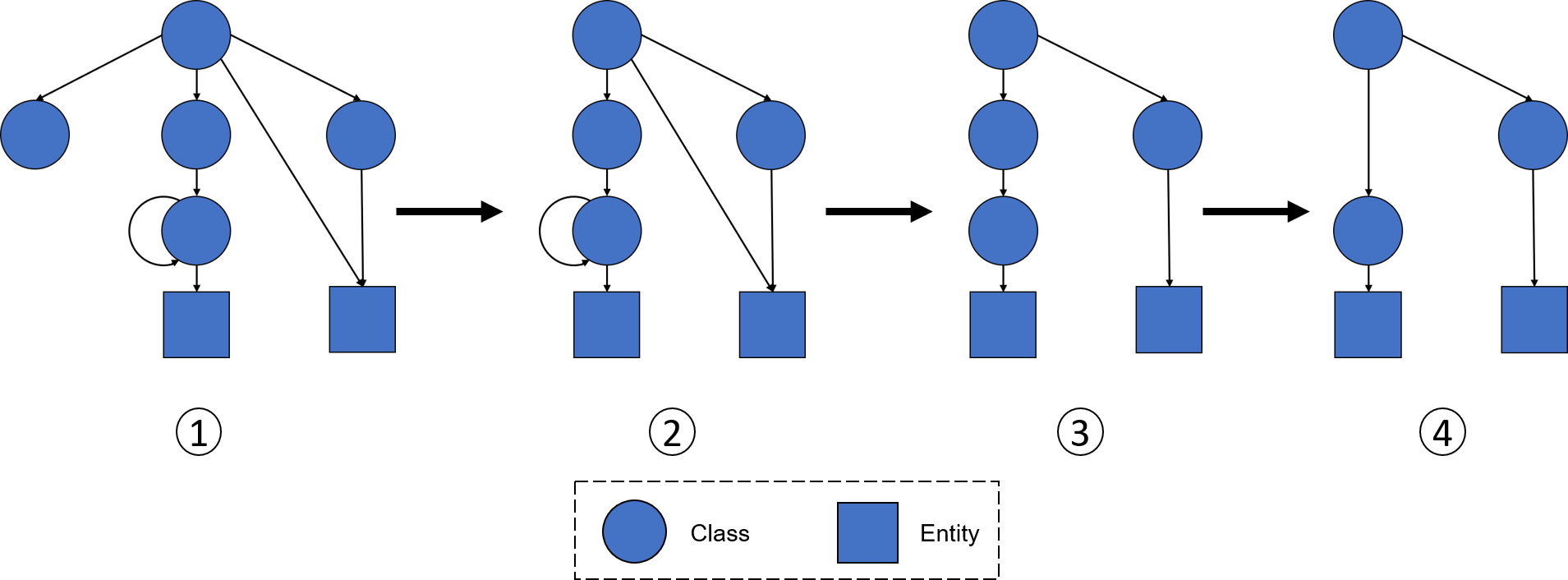}
\caption{Overview of the steps for the creation of the DAG.} \label{fig2}
\end{figure}

\label{sec:method}
\subsubsection{Subgraph Generation}
\label{sec:subgraph}
Given the candidate set $C$ for a mention $m$, a directed-acyclic graph (DAG) $G$ is created from the KG, having the general class $Thing$ as its `root' (i.e., the only node without predecessors) and the candidate entities as `leaves' (i.e., the nodes without successors). Note that $G$ cannot be considered a tree as a node can have multiple predecessors (e.g., `Justin Timberlake' is a linked to the class \textit{Musician} and also to the class \textit{Actor}).

First of all, the candidate entities are linked to the classes they belong to (see Figure~\ref{fig2}, step 1). Then,  the classes that are not predecessors of any of the candidates are removed from $G$ (see Figure~\ref{fig2}, step 2). Next, the relations that can be inferred by traversing $G$ through more granular path of relations are also removed, as well as self-pointing relations (see Figure~\ref{fig2}, step 3). Finally, intermediate nodes which only have one direct successor and that successor is not an entity are also iteratively removed from $G$, linking the direct successor to the node's direct predecessors (see Figure~\ref{fig2}, step 4). With this last step, we aim to increase the granularity and ease the disambiguation, as the classes in the higher levels of the hierarchy tend to be more abstract (e.g., for the class \textit{Musician}, the path from the root in DBpedia is \textit{Thing} $\rightarrow$ \textit{Species} $\rightarrow$ \textit{Eukaryote} $\rightarrow$ \textit{Person} $\rightarrow$ \textit{Artist} $\rightarrow$ \textit{Musician}). In some of the more complex KGs (e.g., YAGO), an entity could also be considered a class. Therefore, if there exist other entities in the candidate set that are linked to this entity, an extra preprocessing step is needed to transform the entity into a leaf, by removing the links to its direct successors while linking them to its direct predecessors.

\begin{figure}
\centering
\includegraphics[width=0.8\textwidth]{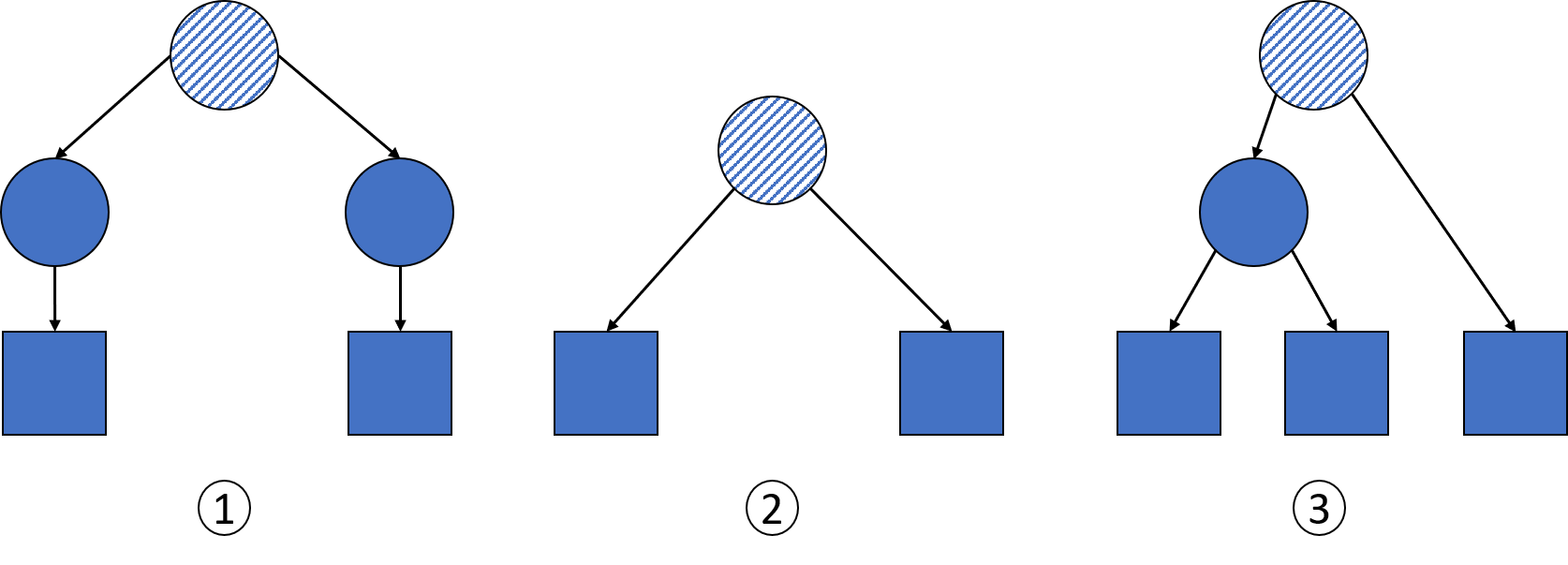}
\caption{Example of the three different configurations of the LCA's direct successors.} \label{fig3}
\end{figure}

\begin{algorithm}[H]
\small
\caption{Pruning candidates}
\label{alg:prune}
\KwIn{Subgraph $G$, mention $m$, document $d$ and entity $descriptions$}
\KwOut{Entity}

\SetKwFunction{LCA}{LCA}
\SetKwFunction{directSuccessors}{directSuccessors}
\SetKwFunction{leaves}{leaves}
\SetKwFunction{multiChoicePrompt}{multiChoicePrompt}
\SetKwFunction{multiChoice}{multiChoice}
\SetKwFunction{prune}{prune}
\SetKwFunction{getClassesAndEntities}{getClassesAndEntities}

candidates $\leftarrow$ \leaves{$G$}\;
\While{$\text{len}(candidates) \neq 1$}{
    $LCA \leftarrow$ \LCA{$G$, candidates}\;
    $directSuccessors \leftarrow$ \directSuccessors{$G$, $LCA$}\;
    \If{$\text{allDirSuccessorsAreClasses}$}{
        $response \leftarrow$ \multiChoice{$directSuccessors \cup \{None\},m,d$}\;
        \If{$response \neq \text{None}$}{
            $G \leftarrow$ \prune{$G$, $directSuccessors \setminus \{response\}$}\;
        }
        \Else{
            $response \leftarrow$ \multiChoice{$candidates,m,d, descriptions$}\;
            $G \leftarrow$ \prune{$G$, $candidates \setminus \{response\}$}\;
        }
    }
    \ElseIf{$\text{allDirSuccessorsAreEntities}$}{
        $response \leftarrow$ \multiChoice{$directSuccessors,m,d, descriptions$}\;
        $G \leftarrow$ \prune{$G$, $directSuccessors \setminus \{response\}$}\;
    }
    \Else{
        $D_c, D_e \leftarrow$ \getClassesAndEntities{$directSuccessors$}\;
        $response \leftarrow$ \multiChoice{$D_c \cup \{Other\},m,d$}\;
        \If{$response = \text{Other}$}{
            $G \leftarrow$ \prune{$G$, $D_c$}\;
        }
        \Else{
            $G \leftarrow$ \prune{$G$, $directSuccessors \setminus \{response\}$}\;
        }
    }
    candidates $\leftarrow$ \leaves{$G$}\;
}
\end{algorithm}

\subsubsection{Pruning Algorithm}
The pruning algorithm is outlined in Algorithm~\ref{alg:prune}. Given the generated graph $G$ and the initial candidate entities (i.e., its leaves), the algorithm starts by finding the Lowest Common Ancestor (LCA) of the candidate entities. The LCA is defined as the deepest node (i.e., the furthest from the root) which is an ancestor of all the candidates (see dashed nodes in Figure~\ref{fig3}). Then, the direct successors of the LCA are retrieved, which leads to three different scenarios:
\begin{enumerate}
    \item \textbf{All the direct successors are classes (Figure~\ref{fig3}, case 1):} The LLM is prompted to select to which classes the mention $m$ belongs to. All the candidate classes that are not chosen by the LLM are removed from $G$, along with all the nodes that have become disconnected from the root. This case corresponds to lines 5-12 in Algorithm~\ref{alg:prune}.
    \item \textbf{All the direct successors are entities (Figure~\ref{fig3}, case 2):} The LLM is prompted to directly select the entity $m$ refers to. Here, the description of each candidate entity is retrieved from a KB and appended to the prompt. The non-selected candidates are then removed from $G$. This case corresponds to lines 13-15 in Algorithm~\ref{alg:prune}.
    \item \textbf{Direct successors are classes and entities (Figure~\ref{fig3}, case 3):} The direct successors are organized into classes ($D_c$), and entities ($D_e$). The LLM is then prompted to select a class from $D_c' = D_c \cup Other$, where $Other$ is an additional class which encompasses $D_e$. If the LLM selects a class belonging to $D_c$, the remaining classes and the entities $D_e$ are removed from $G$. If $Other$ is selected, the classes from $D_c$ are removed. Finally, the nodes which have become disconnected from the root are also removed. This case corresponds to lines 16-22 in Algorithm~\ref{alg:prune}.
\end{enumerate}
 
During the initial tests it was found that the LLM may not return a valid response when it considered that none of the presented classes matched the mention. Therefore, in \textit{case 1} we additionally add the class \textit{None}, which triggers a \textit{case 2} prompt with the remaining candidates if it is selected. Finally, in order to guarantee that the LLM always has information about the entity before making a decision, the response is assessed by the LLM when a single entity is left after a \textit{case 1} or \textit{case 3} step. If it is negatively evaluated, a complete \textit{case 2} prompt is triggered. 

The algorithm runs until there is only one leaf (i.e., entity) left in $G$, which will be the final response. Therefore, in the worst-case scenario the LLM will be prompted $k$ times.

\section{Experiments}
In this section we discuss the experiments performed. First, we describe the experimental settings, then we evaluate our proposal against different methods and also analyze the effect  of the KG used. Finally we study the different scenarios that lead to our method failing to correctly disambiguate the mention.
\label{sec:experiments}
\subsection{Settings and Datasets}
\label{sec:settings}
% We evaluate our proposal using the same experimental settings as~\cite{chatel24}. This is, we use the same datasets, candidate sets and evaluation metrics in order to produce comparable results.
\subsubsection{Datasets}
We evaluate the approach on ten popular ED datasets, the same as in~\cite{chatel24}, which are from news and online articles (MSN~\cite{msn}, AQU~\cite{aqu}, ACE04~\cite{ace04}, CWEB~\cite{cweb}, R128~\cite{reurss} and R500~\cite{reurss}), from Wikipedia (WIKI~\cite{wiki}, OKE15~\cite{oke15} and OKE16~\cite{oke16}) or from hand-crafted, brief and ambiguous sentences (KORE~\cite{kore50}). These datasets contain documents for which one or various mentions have been annotated with the ground truth entity they refer to. The dataset statistics are summarized in Table~\ref{tab:datastats}.

\begin{table}[htbp]
\centering
\footnotesize
\caption{Overview of ten considered datasets' statistics.}
\label{tab:datastats}
\begin{tabular}{@{}lccc@{}}
\toprule
 & \textbf{\# Docs} & \textbf{\# Mentions} & \textbf{Avg. \# Characters}\\
\midrule
\textbf{KORE} & 50 & 144 & 76.4\\
\textbf{ACE04} & 35 & 257 & 2285.0\\
\textbf{OKE16} & 173 & 288 & 186.2\\
\textbf{R500} & 357 & 524 & 164.8\\
\textbf{OKE15} & 101 & 536 & 183.9\\
\textbf{R128} & 113 & 650 & 818.8\\
\textbf{MSN} & 20 & 656 & 3380.1\\
\textbf{AQU} & 50 & 727 & 1415.9 \\
\textbf{WIKI} & 319 & 6793 & 1624.6\\
\textbf{CWEB} & 320 & 11154 & 7575.9\\
\bottomrule
\end{tabular}
\end{table}
\vspace*{-\baselineskip}
\subsubsection{Candidate Sets} 
To allow comparability, we borrow the candidate sets from~\cite{chatel24}, which combine two methods to obtain sets of size 10. First, as done in previous works~\cite{titov18,genre20,refined22}, Wikipedia hyperlink count statistics from mention-entity pairs are used to generate the candidates. If not enough candidates are found, the set is augmented by generating candidates with the BLINK model~\cite{blink}, which is based on dense retrieval from context and descriptions.
\vspace*{-\baselineskip}
\subsubsection{Knowledge Graphs}
To obtain the hierarchical representation of the classes we use YAGO~\cite{yago}. It primarily leverages the information from Wikipedia's infoboxes for generating the relations between entities, and for the taxonomy it borrows the top-level representation from the schema.org ontology~\cite{schemaorg}, which is further refined by carefully integrating it with the fine-grained Wikidata~\cite{wikidata} taxonomy. Additionally, in Section~\ref{sec:dbvsyago} we study the effect of the granularity of the annotation of the classes. To this end, we use another KG with a more simple class hierarchy, DBpedia~\cite{dbpedia}, which is also built on top of Wikipedia but uses a shallow and manually created ontology to define the representation of classes. Finally, to retrieve the entity descriptions we use Wikipedia as a KB, and they are truncated at 250 characters before being appended to the prompt.
\vspace*{-\baselineskip}
\subsubsection{Evaluation Metric}
We report our results with inKB micro-F1 score (see Equation~\ref{eq:micro}). InKB means that we only consider a mention if the ground truth entity is present in the KG used. To allow comparability between KGs (i.e., YAGO and DBpedia do not have the same entities annotated), we also report the results by considering the percentage of the Gold F1 score achieved. The Gold F1 score is the maximum inKB micro-F1 score that could be obtained, as the candidate sets do not always contain the ground truth entity.  
\begin{equation}
    \text{micro-F1} = \frac{\text{TP}}{\text{TP} + \frac{1}{2}(\text{FP} + \text{FN})}
    \label{eq:micro}
\end{equation}
Additionally, given the differences in dataset sizes we report the weighted average, weighting each score by considering the number of instances of each dataset.
\vspace*{-\baselineskip}
\vspace*{-\baselineskip}
\subsubsection{Large Language Models}
To perform our experiments we use GPT-3.5, concretely the  \textit{gpt-3.5-turbo-1106} model from OpenAI API, setting its temperature to 0 to decrease the randomness and the creativity of the response, as we are interested in factual answers. The reason behind the selection of this LLM is in a trade-off between reasoning capabilities, operating cost and API availability.

\subsection{Results}
\label{sec:results}
To evaluate the proposed method we compare it to a non-enhanced LLM baseline and to ChatEL~\cite{chatel24}, the only approach that to the best of our knowledge also directly prompts LLMs to solve in a zero-shot manner the ED task, without training or fine-tuning any model. Additionally, we compare it to ReFinED~\cite{refined22}, the task-specific model, which requires extensive training, that obtained the best ED performance in the results reported in~\cite{chatel24}:
\begin{itemize}
\item \textbf{Baseline:} The baseline consists in asking the LLM to directly select one of the entities within the set of candidates. Therefore, it does not have the class representation nor the entities’ description. This baseline corresponds to a non-enhanced LLM approach. Its implementation can be found in the Supplemental Material.
\item \textbf{ChatEL~\cite{chatel24}:} The ED task is solved in two steps. First, the LLM is asked to describe what the mention in the document is referring to. Then, another prompt is created asking the LLM to select the candidate entity that best matches the description generated in the previous response, by also enriching the candidates with their descriptions from Wikipedia. It must be noted that in our approach an answer is always returned. However, in~\cite{chatel24} an empty result is produced (i.e., a prediction is not performed) when the LLMs response does not contain any candidate entity, which we observe that happens when the response is not on the candidate set or when there is not enough context. This affects the computation of the precision (and thus the inKB and Gold F1-score), as the number of false negatives can potentially be reduced. These observed differences in the F1-score have been mitigated by computing the achieved gold percentage, making the proposals comparable.
\item \textbf{ReFinED~\cite{refined22}:} Is a ED-specific method built over the RoBERTa architecture, leveraging entity types and descriptions. It is pretrained with a Wikipedia dataset, with more than 100M mention-entity pairs, and fine-tuned on AIDA-CoNLL~\cite{aida}, a news related ED dataset with approximately 25.000 annotated mentions. 
\end{itemize}

The results are shown in Table~\ref{tab:results}. First of all, it can be observed that the proposed approach outperforms the baseline in all of the datasets. This demonstrates that even with the vast amounts of data with which the LLMs have been trained and their reasoning capabilities, the addition of external knowledge on the prompts and the guidance during the disambiguation can be helpful to improve the performance on the ED task. One of the most frequent mistakes made by the baseline approach is to give more importance to the context than to the mention. For instance, in the sentence \textit{`A six-game begins this Friday in \underline{Phoenix} and the team hopes to get O'Neal [...]’}, the baseline links the mention to the entity `Phoenix Suns’, presumably given the basketball context. However, the mention is referring to a place, which is correctly resolved by the KG-enhanced approach, as in the first iteration the LLM correctly disambiguates between the classes \textit{Organization}, \textit{Place}, \textit{Product} or \textit{FictionalEntity}.

\begin{sidewaystable}[htbp]
\centering
\caption{Results for the inKB micro F1-score for the ED experiments with ten datasets. The ChatEL and ReFinED scores are taken from the results reported by the authors in~\cite{chatel24}. The weighted average weights each score by taking into consideration the sizes of the datasets. The best score for each dataset is highlighted in \textbf{bold}.}
\label{tab:results}
\setlength{\tabcolsep}{5.5pt}
\renewcommand{\arraystretch}{1.5}
\begin{tabular}{c*{11}{S[table-format=2.1]}|cc}
\toprule
 & & \textbf{KORE} & \textbf{ACE04} & \textbf{OKE16} & \textbf{R500} & \textbf{OKE15} & \textbf{R128} & \textbf{MSN}  & \textbf{AQU}  & \textbf{WIKI} & \textbf{CWEB} & \textbf{Avg.} &\textbf{Wt. avg.}\\
\midrule
\midrule
\multirow{3}{*}{\rotatebox[origin=c]{90}{\textit{Baseline}}} 
& \multicolumn{1}{c}{F1-Score} & 68.2 & 89.1 & 59.0 & 77.4 & 64.1 & 68.7 & 82.3 & 62.4 & 69.5  & 65.0 \\
& \multicolumn{1}{c}{Gold F1} & 76.5 & 95.4 & 82.2 & 85.3 & 82.2 & 83.6 & 94.0 & 96.2 & 89.4 & 89.3 \\
\cmidrule(lr){2-14}
& \multicolumn{1}{c}{\textbf{\% Gold}} & 89.3 & 93.4 & 71.7 & 90.8 & 78.0 & 82.1 & 87.5 & 64.8 &  77.7 & 72.8 & 80.8 & 75.4\\
\midrule
\midrule
\multirow{3}{*}{\rotatebox[origin=c]{90}{\textit{ReFinED\cite{refined22}}}} 
& \multicolumn{1}{c}{F1-Score} & 56.7 & 86.4 & 79.4 & 70.8 & 78.1 & 68.0 & 89.1 & 86.1  & 84.1 & 73.8 \\
& \multicolumn{1}{c}{Gold F1} & 88.0 & 96.9 & 90.3 & 92.1 & 90.3 & 91.1 & 97.0 & 98.1 & 94.4 & 94.3 \\
\cmidrule(lr){2-14}
& \multicolumn{1}{c}{\textbf{\% Gold}} & 64.4 & 89.1 & \textbf{87.9} & 76.8 & 86.4 & 74.6 & \textbf{91.8} & \textbf{87.7} & \textbf{89.0} & \textbf{78.2} & 82.8 & \textbf{82.6} \\
\midrule
\midrule
\multirow{3}{*}{\rotatebox[origin=c]{90}{\textit{ChatEL\cite{chatel24}}}} 
& \multicolumn{1}{c}{F1-Score} & 78.7 & 89.3 & 75.2 & 82.2 & 75.8 & 78.9 & 88.1 & 76.7  & 79.1 & 70.9 \\
& \multicolumn{1}{c}{Gold F1} & 88.0 & 96.9 & 90.3 & 92.1 & 90.3 & 91.1 & 97.0 & 98.1 & 94.4 & 94.3 \\
\cmidrule(lr){2-14}
& \multicolumn{1}{c}{\textbf{\% Gold}} & 89.4 & 92.1 & 83.2 & 89.2 & 83.9 & 86.6 & 90.8 & 78.1 & 83.7 & 75.1 & 85.2 & 79.3 \\
\midrule
\midrule
\multirow{3}{*}{\rotatebox[origin=c]{90}{\textit{$\textit{Our}_{\textit{DBpedia}}$}}} 
& \multicolumn{1}{c}{F1-Score} & 71.3 & 89.4 & 65.9 & 75.4 & 73.5 & 75.0 & 84.2 & 72.0 & 72.5  & 67.7\\
& \multicolumn{1}{c}{Gold F1} & 80.1 & 95.7 & 79.8 & 85.6 & 82.2 & 85.9 & 94.1 & 96.3 & 90.4 & 89.4 \\
\cmidrule(lr){2-14}
& \multicolumn{1}{c}{\textbf{\% Gold}} & 88.9 & 93.3 & 82.5 & 88.0 & \textbf{89.3} & 87.3 & 89.4 & 74.8 & 80.2 & 75.7 & 85.0 & 78.8\\
\midrule
\midrule
\multirow{3}{*}{\rotatebox[origin=c]{90}{$\textit{Our}_{\textit{YAGO}}$}}
& \multicolumn{1}{c}{F1-Score} & 71.8 & 88.7 & 65.8 & 78.3 & 70.3 & 75.8 & 81.2 & 72.0 &  74.4 & 69.6 \\
& \multicolumn{1}{c}{Gold F1} & 79.6 & 94.3 & 83.7 & 85.2 & 82.3 & 84.8 & 92.2 & 94.4 & 88.9 & 89.3\\
\cmidrule(lr){2-14}
& \multicolumn{1}{c}{\textbf{\% Gold}} & \textbf{90.1} & \textbf{94.0} & 78.6 & \textbf{91.9} & 85.4 & \textbf{89.4} & 88.0 & 76.2 & 83.6 & 77.9 & \textbf{85.5} & 81.1\\
\bottomrule
\end{tabular}
\end{sidewaystable}

Regarding the comparison with ChatEL, it can be observed that better results are obtained by our approach in 6 ouf of 10 datasets, with a weighted average score of 1.8 percentage points higher. Additionally, in the complete ChatEL evaluation GPT-4 is used, which is bigger and more powerful LLM than GPT-3.5~\cite{gpt423} with a cost per token more than 20 times higher.\footnote{https://openai.com/pricing} Therefore, even while using a much less powerful LLM, the proposed approach leads to improvements in the ED task. Additionally, the added cost of the manipulation of the graph structure (e.g., finding the LCA and pruning) is limited by the small number of candidates used in ED, which typically ranges from 5 to 30, and its execution time is two orders of magnitude lower than the LLM calls. 

For the task-specific model, we can observe that it obtains a better performance in 6 of the datasets, and an average weighted  score of 1.5 percentage points higher. However, it is worth noting that it has been trained over a huge Wikipedia dataset and fine-tuned on an ED dataset about news, and for the only dataset out of these domains, KORE, our model outperforms it by more than 25 percentage points. Therefore, the LLM methods show a greater degree of adaptability, and could compensate the decrease in performance on some datasets by not requiring the training of specific models.

\subsection{KG Expressivity Impact}
\label{sec:dbvsyago}

\begin{figure}
\includegraphics[width=0.9\textwidth]{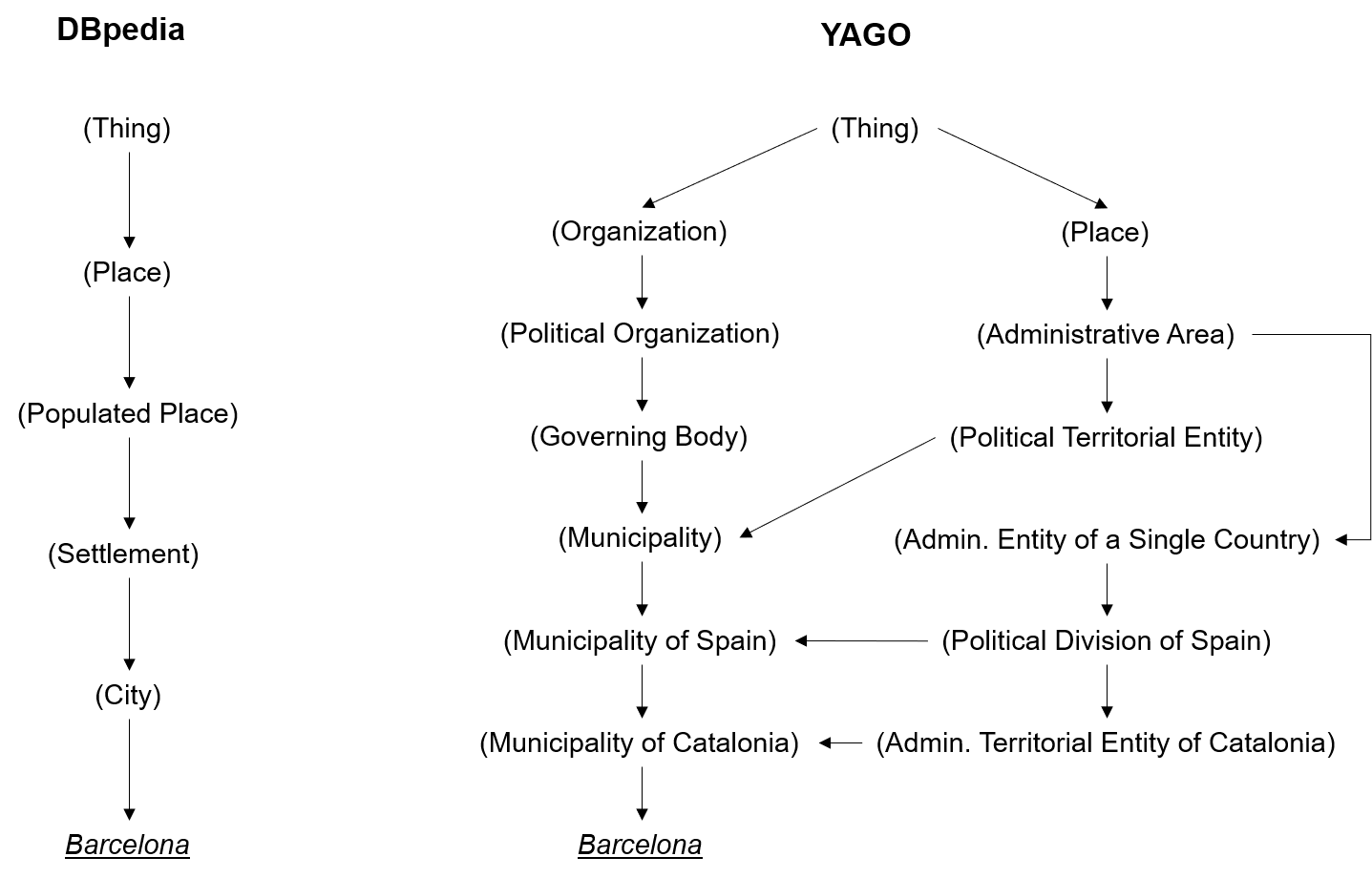}
\caption{Class representation of the entity Barcelona in DBpedia (left) and YAGO (right) KGs.} \label{fig4}
\end{figure}

In this section we evaluate how the differences in the semantic expressivity of the taxonomy of classes in the KG affects our approach. Concretely, we explore if reducing the granularity of the taxonomy affects its disambiguation capabilities. To this end, we use YAGO and DBpedia KGs, whose statistics are summarized in Table~\ref{tab:kgstats}. It can be observed that YAGO has more than a thousand times as many classes as DBpedia, and nearly doubles the average depth of the path from an entity to the root. Therefore, YAGO has a more granular class representation and also annotates more semantic interpretations of the entities. For instance, as it is exemplified in Figure~\ref{fig4}, in the annotation of Barcelona in YAGO a distinction is made between its representation as a \textit{Place}  and as a \textit{Organization}, whereas in DBpedia Barcelona is only considered as a \textit{Place}. Additionally, we can also observe the difference in the number of classes and its granularity. For example, DBpedia stops at the city level, while YAGO classifies the municipalities also within their country and region.

To evaluate the two KGs under study, we repeat the same experimental settings as in Section~\ref{sec:settings}, keeping in mind that the inKB entities do not completely overlap on both KGs, thus affecting the Gold F1-score. The results can be seen in Table~\ref{tab:results}, where in 7 out of 10 datasets the more granular class representation, YAGO, has a better performance, and the weighted average score is 2.4 percentage points higher. This reinforces the hypothesis that having a more semantically rich class taxonomy can help in the disambiguation task. We can observe that YAGO does not outperform DBpedia primarily in the OKE datasets, which contain a large number of mentions referring to generic occupations (e.g., Governor, Judge, Engineer, etc.). For instance, in the second iteration of the method for the sentence \textit{`As \underline{governor}, Reagan raised taxes [...]’}, the disambiguation is between  the entity `Governor' which is the ground truth answer, and the class \textit{Head of Government}, which has other entities as successors (e.g., `Governor of California'). This causes a \textit{case 3} (see Section~\ref{sec:method}) disambiguation between \textit{Head of Government} and \textit{Other}, which leads to the LLM selecting the former as it properly fits the context.

\begin{table}[htbp]
\footnotesize
\centering
\caption{Metric comparisons from YAGO and DBpedia KGs~\cite{kgweb}.}
\label{tab:kgstats}
\begin{tabular}{lS[table-format=7.0]S[table-format=7.0]}
\toprule
 & {\textbf{DBpedia}} & {\textbf{YAGO}} \\
\midrule
\# Instances          & {5,044,223} & {6,349,359} \\
\# Classes            & {760}       & {819,292}   \\
Avg. tree depth       & {3.51}      & {6.61}      \\
Avg. branching factor & {4.53}      & {8.48}      \\
\bottomrule
\end{tabular}
\end{table}

Regarding the number of iterations both KGs exhibit a similar behavior, having a mean value close to 2.2 (see Table~\ref{tab:iterations}). Therefore, even though YAGO has a deeper taxonomy, it is compensated by its superiority in semantic expressivity and mitigated by the elimination of intermediary nodes in the preprocessing step (see Section~\ref{sec:subgraph}).
Hence, given that the execution time of the graph manipulation is two orders of magnitude lower than the LLM calls, using deeper graphs does not significantly affect the performance.

\begin{table}[htbp]
\footnotesize
\centering 
\caption{Percentage of disambiguated entities in which the pruning algorithm reached a final single entity within the specified number of iterations.}
\label{tab:iterations}
\begin{tabular}{l *{6}{ @{\hspace{5mm}}S[table-format=2.2]}  @{\hspace{5mm}}  | S[table-format=1.2]}
\toprule
 & \multicolumn{6}{c}{\textbf{Iterations}} & {\textbf{Avg. Iterations}} \\ 
\cmidrule(lr){2-7} 
 & \textbf{1} & \textbf{2} & \textbf{3} & \textbf{4} & \textbf{5} & \textbf{6} & {\textbf{}} \\ 
\midrule
\textbf{YAGO}    & 26.24\%    & 37.36\%    & 26.60\%    & 8.30\%     & 1.32\%     & 0.15\%     & 2.21          \\
\textbf{DBpedia} & 23.57\%    & 43.00\%    & 26.68\%    & 6.12\%     & 0.42\%     & 0.01\%     & 2.18   \\
\bottomrule
\end{tabular}
\end{table}

\subsection{Error Analysis}
\label{sec:error-analysis}
We thoroughly examined and categorized the scenarios that led to our method producing an incorrect disambiguation, as understanding them is crucial for assessing the capabilities and limitations of LLMs in this task.
\vspace*{-\baselineskip}
\subsubsection{Ground truth errors}
These errors consider the inaccuracies in the annotation of the datasets. For instance, in the sentence \textit{`[...] it is required excellent \underline{English} communication skills [...]'}, the mention English is annotated as `England' instead of `English Language'.
\vspace*{-\baselineskip}
\subsubsection{KG errors}
These errors encompass the problems derived from the annotation of the entities’ classes in the KGs. For example, in the sentence \textit{`Mars, Galaxy and \underline{Bounty} are chocolate [...]'} the ground truth answer `Bounty (chocolate bar)' is wrongly annotated in DBpedia as an \textit{Architectural Structure}, causing the pruning algorithm to fail. Additionally, the annotations could also suffer from inconsistencies. For instance, the `Supreme Court of Florida' falls under the \textit{Organization} class, while the `Supreme Court of California' is considered a \textit{Building}. 
\vspace*{-\baselineskip}
\subsubsection{Ambiguous errors} 
Some datasets contain sentences with high degree of ambiguity. For instance, in the sentence \textit{`\underline{Justin}, Stefani and Kate are among the most popular people both on MTV and Twitter'}, the disambiguation between `Justin Timberlake' and `Justin Bieber' is not clear as both are popular celebrities in those platforms and have collaborated with the other mentioned artists. Moreover, there are some ground truth labels that could be argued to be incorrect. For example, in the sentence \textit{'accepted the post of \underline{principal} and only teacher at a primary school in rural Blaauwbosch, Newcastle.'}, principal is annotated in the ground truth as `Principal (Academia)', yet for primary schools in the UK a more appropriate term would be `head teacher', which is also found in the candidate set. 
\vspace*{-\baselineskip}
\subsubsection{LLM errors}
Finally, some errors are produced by the LLM's response. These are usually originated by the LLM missing information from the context and incorrectly resolving the entity or by wrongly interpreting the mention and assigning it to an erroneous class.

\vspace{\baselineskip}
\noindent In Table~\ref{tab:errors}, all the errors from the two smaller datasets (i.e., KORE and ACE2004) have been classified according to the presented types of error. This study has not been extended to all the datasets as it is unfeasible due to their sizes. Regarding the ground truth error, it corresponds to the sentence \textit{`\underline{Onassis} married Kennedy on October 20, 1968'}, where the mention Onassis is annotated as `Jacqueline Kennedy Onassis' instead of `Aristotle Onassis'. For the KG errors, 2 are originated from a missing class annotation and 1 from a wrong labeling of an entity. Also, 3 errors for the ambiguous sentences are originated by the context not being sufficient to disambiguate the mention and in 2 of them the LLM’s response could arguably be considered also correct (e.g, in the sentence \textit{`The \underline{Isle of Wight festival} in 1970 was the biggest at its time'}, the mention could be both referring to the musical festival and to the concrete festival's edition). Finally, 5 of the LLM errors are caused by missed context (e.g., in the short sentence \textit{`Tiger lost the \underline{US Open}’}, the mention Tiger, likely referring to Tiger Woods, helps to disambiguate between `US Open (tennis)' and `US Open (golf)' but it is missed by the LLM) and 7 by a wrong interpretation of the class (e.g., in the sentence \textit{`[...] ran adjacent to an advertisement for a golf tournament on \underline{Fox Sports} sponsored by Sun Microsystems.’} the mention is interpreted as a TV program rather than a TV channel). 

These last LLM errors could potentially be solved by using LLMs with more powerful reasoning capabilities. To explore this idea, a small experiment with GPT-4 and Mistral Large~\cite{mistral-large} has been run, where the models are able to correctly disambiguate 8 and 7 of these 12 errors, respectively. 

\begin{table}[htbp]
\vspace*{-\baselineskip}
\centering
\small
\caption{Error types for the ACE2004 and KORE datasets, using YAGO as the KG.}
\label{tab:errors}
\begin{tabular}{@{}cc@{}}
\toprule
\textbf{Error Type} & \textbf{\# Errors}\\
\midrule
LLM error & 12\\
Ambiguous error & 5\\
KG error & 3\\
Ground truth error & 1\\
\bottomrule
\end{tabular}
\end{table}
\vspace*{-\baselineskip}
\vspace*{-\baselineskip}
\section{Conclusions}
\label{sec:conclusions}
In this work we present a novel method to enhance LLMs with KGs to solve the ED task. For this purpose, we leverage the entities’ class taxonomy annotated in a KG to gradually prune the candidates’ search space. Additionally, when the disambiguation is at the entity level we add the entities’ descriptions to the prompt. This proposal allows solving the ED task without training task-specific models or fine-tuning them on domain-specific or new data, which is a time-consuming and expensive process. In the experiments we show that the proposed method outperforms both non-enhanced and description-enhanced LLM approaches, and that it has a higher degree of adaptability to different domains than task-specific methods, which rely on the data they have been trained with. Additionally, we observe how using more semantically expressive KGs improves the ED results without degrading the pruning algorithm’s performance. Finally, we analyze the different disambiguation errors and classify them according to their type, drawing conclusions about them. Specifically, and as a future line of work, the usage of more powerful LLMs could be studied, which could help in the disambiguation of difficult mentions.

\paragraph*{Supplemental Material Statement:} Datasets and scripts containing the prompts and the algorithms can be found in the attached repository.\footnote{https://github.com/doubleBlindReview2048/KGLLMs4ED}

\paragraph*{Disclaimer:} This preprint has not undergone peer review (when applicable) or any post-submission improvements or corrections. The Version of Record  of this contribution is published in The Semantic Web – ISWC 2024: 23rd International Semantic Web Conference, Baltimore, MD, USA, November 11–15, 2024, Proceedings, Part I, and it is available online at https://doi.org/10.1007/978-3-031-77844-5\_9 . 
\bibliographystyle{splncs04}

\end{document}